\documentclass[11pt,a4paper]{article}
\usepackage[hyperref]{acl2020}
\usepackage{times}
\usepackage{latexsym}

\usepackage{microtype}
\aclfinalcopy 
\def\aclpaperid{147} 

\usepackage{CJKutf8}
\usepackage{amsmath,amsfonts}
\usepackage{booktabs}
\usepackage{multirow}
\usepackage{graphics}
\usepackage{booktabs}
\usepackage{graphicx}
\usepackage{subcaption}

\title{SpellGCN: Incorporating Phonological and Visual Similarities into Language Models for Chinese Spelling Check}

\author{Xingyi Cheng\thanks{~~Equal contribution.}  \quad Weidi Xu$^{*}$ \quad Kunlong Chen \\ { \bf  Shaohua Jiang \quad Feng Wang \quad Taifeng Wang \quad Wei Chu  \quad Yuan Qi } \\
Ant Financial Services Group \\
{\tt \{fanyin.cxy,weidi.xwd,kunlong.ckl,shaohua.jsh,zifan.wf,} \\
{\tt taifeng.wang,weichu.cw,yuan.qi\}@alibaba-inc.com}}

\date{}

\begin{document}
\begin{CJK*}{UTF8}{gbsn}
\maketitle
\begin{abstract}
Chinese Spelling Check (CSC) is a task to detect and correct spelling errors in Chinese natural language.
Existing methods have made attempts to incorporate the similarity knowledge between Chinese characters.
However, they take the similarity knowledge as either an external input resource or just heuristic rules.
This paper proposes to incorporate phonological and visual similarity knowledge into language models for CSC via a specialized graph convolutional network (SpellGCN).
The model builds a graph over the characters, and SpellGCN is learned to map this graph into a set of inter-dependent character classifiers. 
These classifiers are applied to the representations extracted by another network, such as BERT, enabling the whole network to be end-to-end trainable.
Experiments~\footnote{The dataset and all code for this paper is available at https://github.com/ACL2020SpellGCN/SpellGCN
} are conducted on three human-annotated datasets.
Our method achieves superior performance against previous models by a large margin. 
\end{abstract}

\section{Introduction}

Spelling errors are common in our daily life, caused typically by human writing, automatic speech recognition,  and optical character recognition systems.
Among these errors, misspelling a character frequently occurs due to the similarity between characters. 
In Chinese, many characters are phonologically and visually similar, but semantically very different.
According to ~\citet{liu10}, about 83\% of errors are related to phonological similarity and 48\% are related to visual similarity.
The Chinese Spelling Check (CSC) task aims to detect and correct such misuse of the Chinese language.
Despite recent development, CSC remains a challenging task.
Notably, the spelling checking on Chinese is very different from English, due to its language nature.
Chinese is a language consisting of many pictographic characters without word delimiters.
And the meaning of each character changes dramatically when the context changes.
Therefore, a CSC system needs to recognize the semantics and aggregate the surrounding information for necessary modifications.

\begin{table}
\small
\begin{tabular}{c|c}
\toprule
Input& 餐厅的{\color{orange}换经费产}适合约会 \\
(phonics)&c\=an t\=ing d\=e {\color{orange}hu\`an j\=ing f\`ei ch\'an} sh\`i h\'e yu\=e hu\`i \\
\hline
BERT & 餐厅的{\color{orange}月消费最}适合约会 \\
(phonics)&c\=an t\=ing d\=e {\color{orange}yu\`e xi\=ao f\`ei zu\`i} sh\`i h\'e yu\=e hu\`i \\
\hline 
+SpellGCN &餐厅的{\color{blue}环境非常}适合约会\\
(phonics)&c\=an t\=ing d\=e {\color{blue}hu\'an j\`ing f\=ei ch\'ang} sh\`i h\'e yu\=e hu\`i\\
\bottomrule
\end{tabular}
\caption{A CSC data sample from SIGHAN 2014~\cite{DBLP:conf/acl-sighan/YuLTC14} with ID B1-3440-2, the {\color{orange}incorrect}/{\color{blue}correct} characters are in {\color{orange}orange}/{\color{blue}blue}.
A BERT model modifies the text into a sentence that is semantically reasonable but dissimilar in pronunciation.
By incorporating both phonological and visual similarities, our new method SpellGCN can generate a sentence that is both semantically sensible and phonically similar to the original sentence.
The sentence output from SpellGCN means ``this restaurant is very suitable for dating''.
}
\label{tab:example}
\end{table}

Previous methods followed the line of generative models. 
They used either language models~\cite{cscsmt13,liu10,heuristic14} or  sequence-to-sequence models~\cite{pn19}.
To fuse the external knowledge of the similarity between characters, some of them leveraged a confusion set, which contains a set of similar character pairs.
For instance, ~\citet{heuristic14} proposed to produce several candidates by retrieving the confusion set and then filter them via language models.
~\citet{pn19} used a pointer network to copy a similar character from the confusion set.
These methods attempted to utilize the similarity information to confine the candidates, rather than modeling the relationship between characters explicitly.

In this paper, we propose a novel spelling check convolutional graph network (SpellGCN) that captures the pronunciation/shape similarity and explore the prior dependencies between characters.
Specifically, two similarity graphs are constructed for the pronunciation and shape relationship correspondingly.
SpellGCN takes the graphs as input and generates for each character a vector representation after the interaction between similar characters.
These representations are then constructed into a character classifier for the semantic representation extracted from another backbone module.
We use BERT~\cite{bert19} due to its powerful semantic capacity.
Combining the graph representations with BERT, SpellGCN can leverage the similarity knowledge and generate the right corrections accordingly.
Regarding the example as in Table~\ref{tab:example}, SpellGCN is able to modify the sentence correctly within the pronunciation constraint.


Experiments were conducted on three open benchmarks.
The results demonstrate that SpellGCN improves BERT evidently, outperforming all competitor models by a large margin.

In summary, our contributions are as follows:
\begin{itemize}
\item We propose a novel end-to-end trainable SpellGCN to integrate the pronunciation and shape similarities into the semantic space.
Its essential components such as the specialized graph convolution and attentive combination operations are carefully investigated.
\item We investigate the performance of SpellGCN both quantitatively and qualitatively. 
Experimental results indicate that our method achieves the best results on three benchmark datasets.
\end{itemize}

\section{Related Work}
The CSC task is a long-standing problem and has attracted much attention from the community.
The research emerges in recent years ~\cite{DBLP:conf/acl-sighan/JiaWZ13,sighan14,heuristic14,sighan15,DBLP:conf/acl-tea/FungDWLZW17,pn19,hong2019faspell}, together with other topics, e.g., grammar error correction (GEC)~\cite{DBLP:conf/acl-tea/RaoGZX18,DBLP:conf/acl/JiWTGTG17,DBLP:conf/ijcai/ChollampattTN16,DBLP:conf/acl/ZhouWG18}.
CSC focuses on detecting and correcting character errors, while GEC also includes errors that need deletion and insertion.
Previous work handles CSC using unsupervised language models~\cite{cscsmt13,heuristic14}.
The errors are detected/corrected by evaluating the perplexity of sentences/phrases. 
However, these models were unable to condition the correction on the input sentence.
To circumvent this problem, several discriminative sequence tagging methods were adopted for CSC~\cite{DBLP:conf/emnlp/WangSLHZ18}.
For more flexibility and better performance, several sequence-to-sequence models were also employed~\cite{pn19,DBLP:conf/acl/JiWTGTG17,DBLP:conf/ijcai/ChollampattTN16,DBLP:conf/acl/ZhouWG18}, as well as BERT~\cite{hong2019faspell}.

Recent attention was paid to utilizing the external knowledge of character similarity. 
The similarity knowledge can be gathered into a dictionary, i.e., confusion set, where similar pairs are stored.
~\citet{heuristic14} first used the dictionary to retrieve similar candidates for potential errors.
~\citet{pn19} incorporated a copy mechanism into a recurrent neural model.
When given similar characters as input, their model uses the copy mechanism to directly copy the character to the target sentence.
In a sense, these models face difficulty in modeling the relationship between similar characters as the similarity information is solely used for candidate selection.
To capture the pronunciation/shape similarity and explore the prior dependencies between characters, we propose to use graph convolution network (GCN)~\cite{gcn17} to model character inter-dependence, which is combined with the pre-training of BERT~\cite{bert19,DBLP:journals/corr/abs-1909-03405} for the CSC task.

GCN has been applied to model the relationship on several tasks.
~\citet{yan2019event} equipped it into the relation extraction task where relations construct a hierarchical tree.
~\citet{DBLP:conf/uic/LiPLXDMWB18,DBLP:conf/ijcnn/ChengZZX18} use it to model spatial-temporal to predict traffic flow.
GCN was also used to model the relationship between labels in a multi-label task~\cite{DBLP:conf/cvpr/ChenWWG19}.
In this paper, it is the first time that GCN is applied successfully into the CSC task.
The relationship in CSC is much different from those tasks where objects in the graph are semantically related.
By contrast, the similar characters are semantically distinct in CSC.
Therefore, we deeply investigate the effect of our SpellGCN and propose several essential techniques.

\section{Approach}
In this section, we elaborate on our method for CSC.
Firstly, the problem formulation is presented.
Then, we introduce the motivations for SpellGCN, followed by its detailed description.
At last, we present its application in the CSC task.

\subsection{Problem Formulation}
The Chinese Spelling Check task aims to detect and correct the errors in the Chinese language.
When given a text sequence $\mathbf{X}=\{x_1, x_2, ..., x_n\}$ consisting of $n$ characters, the model takes $\mathbf{X}$ as input and output a target character sequence $\mathbf{Y}=\{y_1, y_2, ..., y_n\}$.
We formulate the task as a conditional generation problem by modeling and maximizing the conditional probability $p(\mathbf{Y}|\mathbf{X})$.

\subsection{Motivations}

\begin{figure*}
\centering
\includegraphics[width=0.9\textwidth]{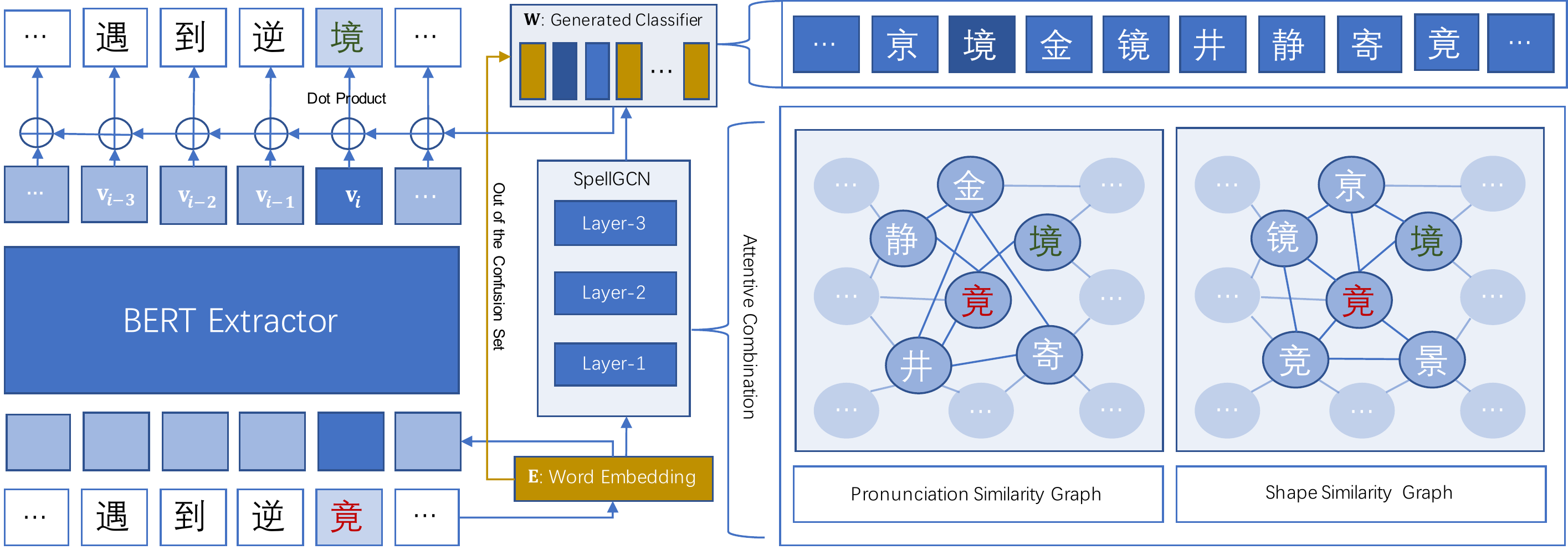}
\caption{The framework of the proposed SpellGCN. 
\textbf{Left}: The characters in the input sentence are processed by the extractor to obtain the semantic representation vectors. 
\textbf{Right}: The phonological or visual similarity knowledge of characters is learned by our SpellGCN. 
Two similarity graphs are used to model the pronunciation and shape similarities respectively, and they are combined via an attentive combination operation.
\textbf{Middle}: The character embedding vectors derived from SpellGCN are used as the target character classifiers
. 
%
}
\label{fig:framework}
\end{figure*}

The framework of the proposed method is depicted in Figure~\ref{fig:framework}.
It consists of two components, i.e., a character representation extractor and a SpellGCN.
The extractor derives a representation vector for each character.
Above the extractor, SpellGCN is used to model the inter-dependence between characters.
It outputs target vectors containing the information of similar characters after interactions.

As illustrated in Table~\ref{tab:example}, a vanilla language model is able to provide feasible corrections in semantic meaning but faces the difficulty in meeting the pronunciation constraint.
Although the correction ``月消费最'' is semantically plausible,
its phonics differs much from ``换经费产'' and ``环境非常''.
This indicates that the similarity information between characters is necessary so that the model can learn to generate related answers.
Previous methods have taken the similarity into consideration.
However, they typically regarded similar characters as potential candidates, neglecting their inter-relationship in terms of pronunciation and shape.
This work makes a preliminary attempt to handle this issue, trying to fuse both the \textit{symbolic space} (phonological and visual similarity knowledge) and the \textit{semantic space} (language semantic knowledge) into one model.
To achieve this, we leverage the power of graph neural network (GNN) to infuse the similarity knowledge directly.
The essential idea is to update the representations by aggregating the information between similar characters.
Intuitively, a model is likely to have a sense of similar symbols when equipped with our method.

Among various GNN models, we use GCN in our implementation.
Since there are up to 5K Chinese characters in the graph, the light-weight GCN is more suitable for our problem.
The proposed SpellGCN is depicted as follows in detail.

\subsection{Structure of SpellGCN}
SpellGCN requires two similarity graphs $\mathbf{A}^p, \mathbf{A}^s$ for pronunciation and shape similarities correspondingly, which are derived from an open-sourced confusion set~\cite{sighan13}. 
For simplicity, the superscript will be omitted if unnecessary and $\textbf{A}$ denotes one of these two similarity graphs.
Each similarity graph is a binary adjacent matrix of size $\mathbb{R}^{N \times N}$, constructed from $N$ characters in the confusion set.
The edge $\mathbf{A}_{i,j} \in \{0,1\}$ between $i$-th character and $j$-th character denotes whether the $(i,j)$ pair exists in the confusion set.

The goal of SpellGCN is to learn a map function to map the input node embedding $\mathbf{H}^l \in \mathbb{R}^{N\times D}$ of $l$-th layer (where $D$ is the dimensionality of character embedding) to a new representation $\mathbf{H}^{l+1}$ via convolutional operation defined by $\mathbf{A}$. 
This map function has two main sub-components: a graph convolution operation and an attentive graph combination operation.

\paragraph{Graph Convolution Operation} 
The graph convolution operation is to absorb the information from neighboring characters in the graph. 
In each layer, the light-weight convolution layer in GCN~\cite{gcn17} is adopted:
\begin{equation}
\begin{aligned}
f(\mathbf{A}, \mathbf{H}^l) &= \hat{\mathbf{A}}\mathbf{H}^l\mathbf{W}_{g}^l \,, \\
\end{aligned}
\end{equation}
where $\mathbf{W}^l_g\in \mathbb{R}^{D\times D}$ is a trainable matrix and $\hat{\mathbf{A}} \in \mathbb{R}^{N\times N}$ is the normalized version of the adjacent matrix $\mathbf{A}$.
For the definition of $\hat{\mathbf{A}}$, we direct you to the original paper~\cite{gcn17}.
Note that we use the character embedding of BERT as the initial node features $\mathbf{H}^0$, and we omit the non-linearity function after convolution.
Since we adopted BERT as our extractor, which has its own learned semantic space, we remove the activation function from the equation to keep the derived representation identical with original space, rather than a completely different space.
During our experiments, using non-linearity activation such as ReLU is ineffective, resulting in a performance drop.

\paragraph{Attentive Graph Combination Operation}
\label{sec:attention}
The graph convolution operation handles the similarity of a single graph.
To combine the pronunciation and shape similarity graphs, the attention mechanism~\cite{DBLP:journals/corr/BahdanauCB14} is adopted.
For each character, we represent the combination operation as follows:
\begin{equation}
\mathbf{C}^l_i=\sum_{k\in \{s, p\}} \alpha^l_{i,k}  f_k(\mathbf{A}^k, \mathbf{H}^l)_{i}\,,
\end{equation}
where $\mathbf{C}^l \in \mathbb{R}^{N\times D}$ and $f_k(\mathbf{A}^k, \mathbf{H}^l)_{i}$ is the $i$-th row of convolved representation of graph $k$, $\alpha_{i,k}$ is a scalar for $i$-th character denoting the weight of graph $k$.
The weight $\alpha_{i,k}$ is computed by 
\begin{equation}
\alpha_{i,k} = \frac{\exp(\mathbf{w}_a f_k(\mathbf{A}^k, \mathbf{H}^l)_{i}/\beta)}{ \sum_{k'} \exp(\mathbf{w}_a  f_{k'}(\mathbf{A}^{k'}, \mathbf{H}^l)_{i}/\beta)},
\end{equation}
where $\mathbf{w}_a \in \mathbb{R}^{D}$ is a learnable vector shared across the layers and $\beta$ is a hyper-parameter which controls the smoothness of attention weights.
We found $\beta$ essential for the attention mechanism.

\paragraph{Accumulated Output}
After graph convolution and attentive combination operations, we obtain a representation $\mathbf{C}^l$ for $l$-th layer.
To maintain the original semantic of the extractor, all outputs of previous layers are accumulated as the output:
\begin{equation}
\mathbf{H}^{l+1} = \mathbf{C}^l + \sum_{i=0}^{l} \mathbf{H}^{i}\,.
\end{equation}
In this way, SpellGCN is able to focus on capturing the knowledge of character similarity, leaving the responsibility of semantic reasoning to the extractor.
Hopefully, each layer can learn to aggregate the information for the specific hop.
During the experiments, the model failed when excluding $\mathbf{H}^0$. 

\subsection{SpellGCN for Chinese Spelling Check}
Here, we introduce how to apply SpellGCN to the CSC task.
Motivated by recent applications of GCN in relationship modeling~\cite{DBLP:conf/cvpr/ChenWWG19,yan2019event}, we use the final output of SpellGCN to be classifiers of the target characters.

\paragraph{Similarity Graphs from Confusion Set}
The similarity graphs used in SpellGCN are constructed from the confusion set provided in ~\cite{sighan13}.
It is a pre-defined set consisting of similar characters for most of ($\sim$95\%) the Chinese characters and these characters are categorized into five categories, i.e., (1) similar shape, (2) same pronunciation and same tone, (3) same pronunciation and different tone, (4) similar pronunciation and same tone, (5) similar pronunciation and different tone.
Since the pronunciation similarity is more fine-grained compared with the shape similarity category, we combine the pronunciation similarities into one graph. 
Consequently, we construct two graphs corresponding to pronunciation and shape similarities. 

\paragraph{Character Representation by Extractor}
The representation of characters used for final classification is given by an extractor.
We can use any model that is able to output representation vectors $\mathbf{V}=\{\mathbf{v}_1, \mathbf{v}_2, ..., \mathbf{v}_n\}$ (where $\mathbf{v}_i \in \mathbb{R}^{D}$) for $n$ characters $\mathbf{X}=\{x_1, x_2, ... ,x_n\}$.
In our experiment, we adopt BERT as the backbone model.
It takes $\mathbf{X}$ as input and uses the output of the last layer as $\mathbf{V}$.
We conduct the experiment using the base version, which has 12 layers, 12 self-attention heads with a hidden size of 768~\footnote{This means $D=$768 in our experiment.}.

\paragraph{SpellGCN as Character Classifier}
When given the representation vector $\mathbf{v}_i$ of a character $x_i$, the model needs to predict a target character through a fully-connected layer whose weight $\mathbf{W} \in \mathbb{R}^{M \times D}$  is configured by the output of SpellGCN ($M$ is the size of the extractor vocabulary):
\begin{equation}
p(\hat{y}_i|\mathbf{X}) = \text{softmax}(\mathbf{W}\mathbf{v}_i) \,.
\end{equation}
Concretely, the output vectors of SpellGCN plays the role of the classifier in our task.
We use the output of the last layer of SpellGCN $\mathbf{H}^L$ (where $L$ is the number of layers) to classify the characters in the confusion set.
And since the confusion set only covers a subset of vocabulary, we use the word embedding of the extractor as the classifier for those excluded by the confusion set.
In this way, denoting $u_i \in \{1,...,N\}$ is the index of confusion set for the $i$-th character in the extractor vocabulary, $\mathbf{W}$ is presented by:
\begin{equation}
\mathbf{W}_i = 
\begin{cases}
\mathbf{H}^L_{u_{i}}, & \text{if } i\text{-th character} \in \text{confusion set}  \\
\mathbf{E}_i, & \text{otherwise} \,,
\end{cases}
\end{equation}
where $\mathbf{E} \in \mathbb{R}^{M\times D}$  is the embedding matrix of extractor.
In brief, we use the embedding from SpellGCN if the character is in the confusion set. Otherwise, the embedding vectors are used as in BERT.
Instead of modeling a large compact graph containing all characters in the extractor vocabulary, we chose this implementation for computational efficiency, since there are around 5K characters in the confusion set and more than 20K characters in the extractor vocabulary.

Overall, the objective is to maximize the log likelihood of target characters:
\begin{equation}
\mathcal{L} = \sum_{\mathbf{X},\mathbf{Y}}\sum_i \log p(\hat{y}_i=y_i|\mathbf{X}) \,.
\end{equation}

\subsection{Prediction Inference}
The CSC task consists of two sub-tasks in evaluation, i.e., detection and correction.
Some previous work~\cite{heuristic14,cscsmt13} used two models for these sub-tasks separately.
In this work, we simply use the character with the highest probability $\arg\max_{\hat{y}_i} p(\hat{y}_i|\mathbf{X})$ as the prediction for the correction task.
And the detection is achieved by checking whether the prediction matches the target character $y_i$.



\section{Experiments}

In this section, we describe our experiment in detail.
We first present the training data and test data, as well as the evaluation metrics.
Then we introduce our main results for SpellGCN.
After that, the ablation studies are made to analyze the effect of the proposed components, followed by a case study.
Finally,  quantitative results are provided.

\subsection{Datasets}
\paragraph{Training Data}
The training data is composed of three training datasets~\cite{sighan13,DBLP:conf/acl-sighan/YuLTC14,sighan15}, which has 10K data samples in total.
Following~\cite{pn19}, we also include additional 271K samples as the training data, which are generated by an automatic method~\cite{DBLP:conf/emnlp/WangSLHZ18}~\footnote{https://github.com/wdimmy/Automatic-Corpus-Generation}.

\paragraph{Test Data}
To evaluate the performance of the proposed method, we used three test datasets from the SIGHAN 2013, SIGHAN 2014, SIGHAN 2015 benchmarks~\cite{sighan13,DBLP:conf/acl-sighan/YuLTC14,sighan15} as in~\cite{pn19}.
We also follow the same data pre-processing procedure, i.e., the characters in these datasets are converted to simplified Chinese using OpenCC~\footnote{https://github.com/BYVoid/}.
The statistic of the data is listed in Table~\ref{tab:stat}.

\begin{table}[]
\small
\centering
\begin{tabular}{@{}l r r r @{}}
\toprule
Training Data                          & \# Line           & Avg. Length   & \# Errors \\ \midrule
\cite{DBLP:conf/emnlp/WangSLHZ18}                    & 271,329            & 44.4          & 382,704   \\
SIGHAN 2013            & 350                & 49.2          & 350       \\
SIGHAN 2014            & 6,526              & 49.7          & 10,087    \\
SIGHAN 2015             & 3,174              & 30.0          & 4,237     \\ \midrule
Total                         & 281,379            & 44.4          & 397,378   \\ \bottomrule \toprule
Test Data                          & \# Line             & Avg. Length   & \# Errors \\ \midrule
SIGHAN 2013             & 1000(1000)                & 74.1          & 1,227     \\
SIGHAN 2014             & 1062(526)                & 50.1          & 782       \\
SIGHAN 2015             & 1100(550)                & 30.5          & 715        \\ \bottomrule \toprule
\multicolumn{2}{l}{Graph}                           & \#  Character & \# Edges  \\ \midrule
\multicolumn{2}{l}{Pronunciation Similarity Graph }  & 4753          & 112,687   \\
\multicolumn{2}{l}{Shape Similarity Graph }         & 4738          & 115,561   \\ \bottomrule
\end{tabular}
\caption{Statistics information of the used data resources. The number in the bracket in \#Line column denotes the number of sentences with errors.}
\label{tab:stat}
\end{table}

\begin{table*}[]
\centering
\resizebox{0.72\textwidth}{!}{%
\begin{tabular}{@{}l|rrr|rrr|rrr|rrr@{}}
\toprule
                                                 & \multicolumn{6}{c|}{Character-level}                                                           & \multicolumn{6}{|c}{Sentence-level}                                                            \\ \midrule
                                                 & \multicolumn{3}{c|}{Detection-level}           & \multicolumn{3}{|c|}{Correction-level}          & \multicolumn{3}{|c|}{Detection-level}           & \multicolumn{3}{|c}{Correction-level}          \\ \midrule
SIGHAN 2013                                      & D-P           & D-R           & D-F           & C-P           & C-R           & C-F           & D-P           & D-R           & D-F           & C-P           & C-R           & C-F           \\ \midrule
LMC~\cite{DBLP:conf/acl-sighan/XieHZHHCH15} & 79.8          & 50.0          & 61.5          & 77.6          & 22.7          & 35.1          & (-)           & (-)           & (-)           & (-)           & (-)           & (-)           \\
SL~\cite{DBLP:conf/emnlp/WangSLHZ18}        & 54.0          & 69.3          & 60.7          & (-)           & (-)           & 52.1          & (-)           & (-)           & (-)           & (-)           & (-)           & (-)           \\
PN~\cite{pn19}                              & 56.8          & 91.4          & 70.1          & 79.7          & 59.4          & 68.1          & (-)           & (-)           & (-)           & (-)           & (-)           & (-)           \\
FASpell~\cite{hong2019faspell}         & (-)           & (-)           & (-)           & (-)           & (-)           & (-)           & 76.2          & 63.2          & 69.1          & 73.1          & 60.5          & 66.2          \\ \midrule
BERT                                              & 80.6         & 88.4          & 84.3           & 98.1          & 87.2          & 92.3          & 79.0           & 72.8          & 75.8          & 77.7          & 71.6          & 74.6 \\
SpellGCN                                         & \textbf{82.6} & \textbf{88.9}  & \textbf{85.7}          & \textbf{98.4}        & \textbf{88.4}  & \textbf{93.1}          & \textbf{80.1}        & \textbf{74.4}  & \textbf{77.2}          & \textbf{78.3}        & \textbf{72.7}  & \textbf{75.4} \\ 
\bottomrule \toprule
SIGHAN 2014                                      & D-P           & D-R           & D-F           & C-P           & C-R           & C-F           & D-P           & D-R           & D-F           & C-P           & C-R           & C-F           \\ \midrule
LMC~\cite{DBLP:conf/acl-sighan/XieHZHHCH15} & 56.4          & 34.8          & 43.0          & 71.1          & 50.2          & 58.8          & (-)           & (-)           & (-)           & (-)           & (-)           & (-)           \\
SL~\cite{DBLP:conf/emnlp/WangSLHZ18}        & 51.9          & 66.2          & 58.2          & (-)           & (-)           & 56.1          & (-)           & (-)           & (-)           & (-)           & (-)           & (-)           \\
PN~\cite{pn19}                              & 63.2          & 82.5          & 71.6          & 79.3          & 68.9          & 73.7          & (-)           & (-)           & (-)           & (-)           & (-)           & (-)           \\
FASpell~\cite{hong2019faspell}         & (-)           & (-)           & (-)           & (-)           & (-)           & (-)           & 61.0            & 53.5          & 57.0          & 59.4          & 52.0          & 55.4          \\ \midrule
BERT                                             & 82.9          & 77.6          & 80.2          & 96.8          & 75.2          & 84.6          & \textbf{65.6}          & 68.1          & 66.8          & \textbf{63.1}          & 65.5          & 64.3          \\
SpellGCN                                         & \textbf{83.6} & \textbf{78.6} & \textbf{81.0} & \textbf{97.2} & \textbf{76.4} & \textbf{85.5} & 65.1 & \textbf{69.5} & \textbf{67.2} & \textbf{63.1} & \textbf{67.2} & \textbf{65.3} \\ 
\bottomrule \toprule
SIGHAN 2015                                      & D-P           & D-R           & D-F           & C-P           & C-R           & C-F           & D-P           & D-R           & D-F           & C-P           & C-R           & C-F           \\ \midrule
LMC~\cite{DBLP:conf/acl-sighan/XieHZHHCH15} & 83.8          & 26.2          & 40.0          & 71.1          & 50.2          & 58.8          & (-)           & (-)           & (-)           & (-)           & (-)           & (-)           \\
SL~\cite{DBLP:conf/emnlp/WangSLHZ18}        & 56.6          & 69.4          & 62.3          & (-)           & (-)           & 57.1          & (-)           & (-)           & (-)           & (-)           & (-)           & (-)           \\
PN~\cite{pn19}                              & 66.8          & 73.1          & 69.8          & 71.5          & 59.5          & 69.9          & (-)           & (-)           & (-)           & (-)           & (-)           & (-)           \\
FASpell~\cite{hong2019faspell}         & (-)           & (-)           & (-)           & (-)           & (-)           & (-)           & 67.6          & 60.0          & 63.5          & 66.6          & 59.1          & 62.6          \\ \midrule
BERT                                             & 87.5          & 85.7          & 86.6          & 95.2          & 81.5          & 87.8          & 73.7          & 78.2          & 75.9          & 70.9          & 75.2          & 73.0          \\
SpellGCN                                         & \textbf{88.9}          & \textbf{87.7} & \textbf{88.3} & \textbf{95.7}          & \textbf{83.9} & \textbf{89.4} & \textbf{74.8}          & \textbf{80.7} & \textbf{77.7}          & \textbf{72.1}    & \textbf{77.7} & \textbf{75.9}          \\ \bottomrule
\end{tabular}%
}
\caption{The performance of our method and baseline models (\%). 
D, C denote the detection, correction, respectively. 
P, R, F denote the precision, recall and F1 score, respectively. 
The results of BERT are from our own implementation.
Best results are in \textbf{bold}.
We performed additional fine-tuning on SIGHAN13 for 6 epochs as the data distribution in SIGHAN13 differs from other datasets, e.g. ``的'', ``得'' and ``地'' are rarely distinguished.
}
\label{tab:mainresults}
\end{table*}

\paragraph{Baseline Models}
We compare our method with five typical baselines.
\begin{itemize}
\item LMC~\cite{DBLP:conf/acl-sighan/XieHZHHCH15}: This method utilizes the confusion set to replace the characters and then evaluates the modified sentence via a N-gram \textbf{L}anguage \textbf{M}odel.
\item SL~\cite{DBLP:conf/emnlp/WangSLHZ18}: This method proposes a pipeline where a \textbf{S}equence \textbf{L}abeling model is adopted for detection. The incorrect characters are marked as 1 (0 otherwise).
\item PN~\cite{pn19}: This method incorporates a \textbf{P}ointer \textbf{N}etwork to consider the extra candidates from the confusion set.
\item FASpell~\cite{hong2019faspell}: This model utilizes a specialized candidate selection method based on the similarity metric. This metric is measured using some empirical methods, e.g., edit distance, rather than a pre-defined confusion set. 
\item BERT~\cite{bert19}: The word embedding is used as the softmax layer on the top of BERT for the CSC task. We trained this model using the same setting, i.e., the comparable model w/o SpellGCN.
\end{itemize}

\paragraph{Evaluation Metrics}
The precision, recall and F1 scores are reported as the evaluation metrics, which are commonly used in the CSC tasks.
These metrics are provided for the detection and correction sub-tasks.
Besides the evaluation on the character level, we also report the sentence-level metrics on the detection and correction sub-tasks, which is more appealing for real-world applications.
On the sentence level, we consider a sentence to be correctly annotated only if all errors in the sentence are corrected as in~\cite{hong2019faspell}~\footnote{https://github.com/iqiyi/FASPell}. 
On the character level, we calculate the metrics using the evaluation script from ~\cite{pn19}~\footnote{https://github.com/wdimmy/Confusionset-guided-Pointer-Networks-for-Chinese-Spelling-Check}.
We also evaluated BERT and SpellGCN by the official evaluation metrics tools\footnote{http://nlp.ee.ncu.edu.tw/resource/csc.html}, which gives False Positive Rate~(FTR), Accuracy and Precision/Recall/F1.

\subsection{Hyper-parameters}
Our code is based on the repository of BERT~\footnote{https://github.com/google-research/bert}.
We fine-tune the models using AdamW~\cite{loshchilov2018decoupled} optimizer for 6 epochs with a batch size of 32 and a learning rate of 5e-5.
The number of the layer in SpellGCN is 2, and the attentive combination operation with factor $3$ is used.
All experiments were conducted for 4 runs and the averaged metric is reported.
The code and trained models will be released publicly after review (currently, the code is attached in the supplementary files).

\begin{table*}[]
\centering
\resizebox{0.60\textwidth}{!}{%
\begin{tabular}{@{}l |r |r |r |r r r |r r r@{}}
\toprule
SIGHAN 2014                          & FPR & D-A & C-A & D-P & D-R & D-F & C-P & C-R & C-F 
                          \\ \midrule
BERT                     & 15.3            & 76.8          & 75.7 & 81.9 & 68.9 & 74.9 & 81.4 & 66.7 & 73.3   \\
SpellGCN            & \textbf{14.1}              & \textbf{77.7}          & \textbf{76.9} & \textbf{83.1} & \textbf{69.5} & \textbf{75.7} & \textbf{82.8} & \textbf{67.8} & \textbf{74.5}    \\ \midrule
SIGHAN 2015                          & FPR & D-A & C-A & D-P & D-R & D-F & C-P & C-R & C-F \\ \midrule
BERT                     & 13.6            & 83.0 & 81.5 & 85.9 & 78.9 & 82.3 & \textbf{85.5} & 75.8 & 80.5    \\ 
SpellGCN            & \textbf{13.2}              & \textbf{83.7}  & \textbf{82.2} & \textbf{85.9} & \textbf{80.6} & \textbf{83.1} & 85.4 & \textbf{77.6} & \textbf{81.3}     \\
\bottomrule

\end{tabular}
}
\caption{The performance of BERT and SpellGCN evaluated by official tools on SIGHAN 2014 and SIGHAN 2015. FPR denotes the false positive rate and A denotes the accuracy. D-A and C-A denote detection accuracy and correction accuracy.}
\label{tab:off_eval}
\end{table*}

\subsection{Main Results}

\begin{figure}
\centering
\includegraphics[width=.9\linewidth,height=0.6\linewidth]{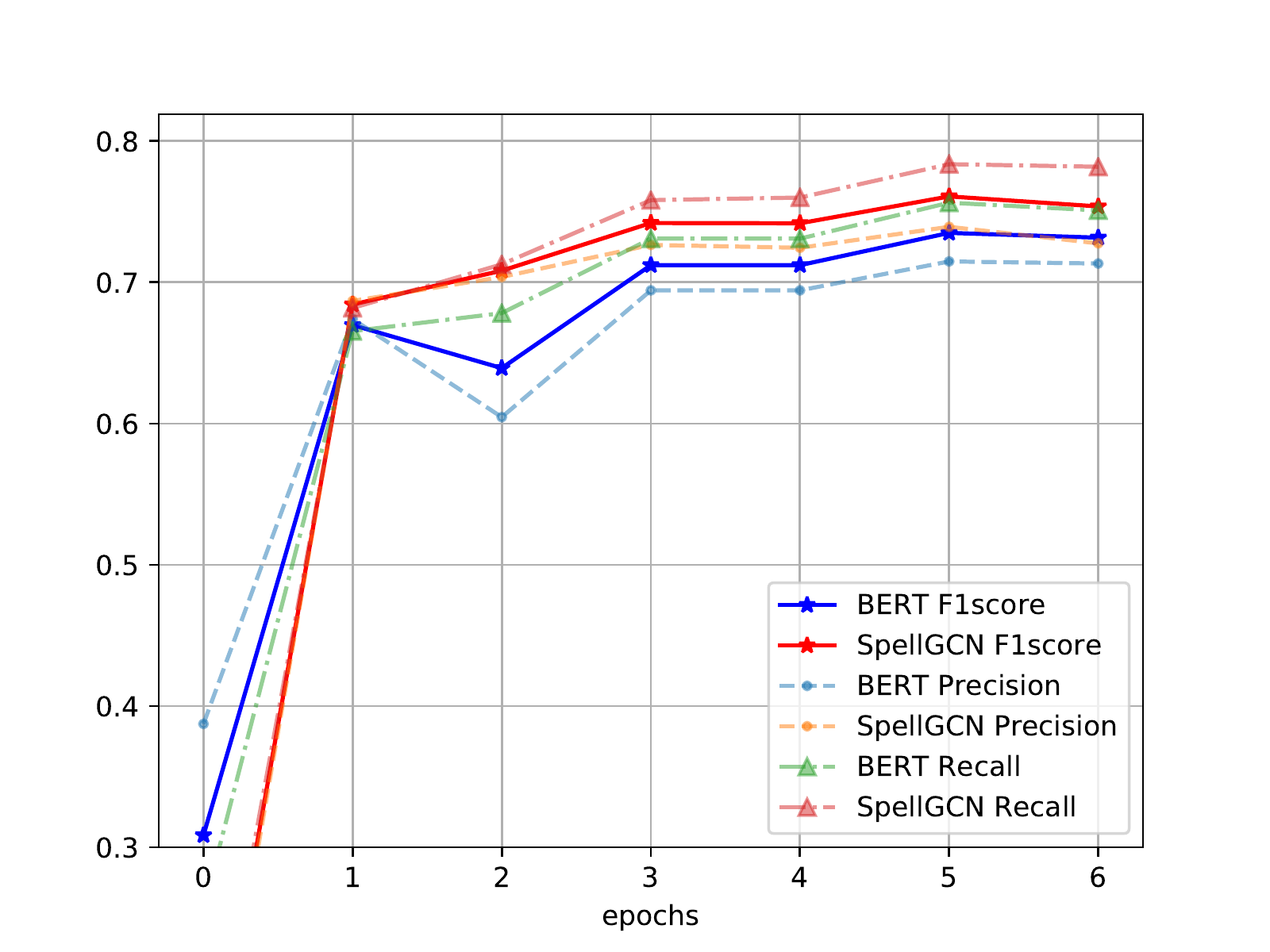}
\caption{The test curves for sentence-level correction metrics with and without SpellGCN w.r.t. the number of training epochs on SIGHAN 2015.}
\label{fig:test_curve}
\end{figure}

Table~\ref{tab:mainresults} shows the performance of the proposed method on the three CSC datasets, compared with five typical CSC systems.
When using SpellGCN, the model achieves better results in all test sets against vanilla BERT, which verifies its effectiveness.
The improvement is considerable with such a large amount of training data (cf. the comparison in Figure~\ref{fig:test_curve}).
This indicates the similarity knowledge is essential for CSC and it can hardly be learned by simply increasing the data amount.
In terms of sentence-level F1score metric in the correction sub-task, i.e., C-F score in the last column, the improvements against previous best results (FASPell) are 9.2\%, 9.7\% and 13.3\% respectively.
Nevertheless, it should be noted that FASpell was trained on different training data while this paper follows the setting mentioned in the PN paper~\cite{pn19}.
Ideally, our method is compatible with FASpell and better results can be achieved when FASpell is employed.

FASpell used their own metrics, which are different from the sentence-level false postive and false negtivate counting strategy of the official evaluation toolkit.
We used the scripts by PGNet and FASpell to compute their metrics for fair comparison.  
We further add the official evaluation results of BERT and SpellGCN in Table~\ref{tab:off_eval}. 
Actually, SpellGCN consistently improves the performance when evaluated by the PGNet/FASpell scripts and the official evaluation toolkit. We will add the FPR results in our revision. The FPR scores are 14.1\% (SpellGCN) v.s. 15.3\% (BERT) on SIGHAN 14, and 13.2\% (SpellGCN) v.s. 13.6\% (BERT) on SIGHAN 15. FPR on SIGHAN 13 is statistically meaningless since almost all the tested sentences have the spelling errors.



\subsection{Ablation Studies}
In this subsection, we analyze the effect of several components, including the number of layers and the attention mechanism.
The ablation experiments were performed using 10K training data.

\begin{figure}
\centering
\includegraphics[width=0.4\textwidth,height=0.6\linewidth]{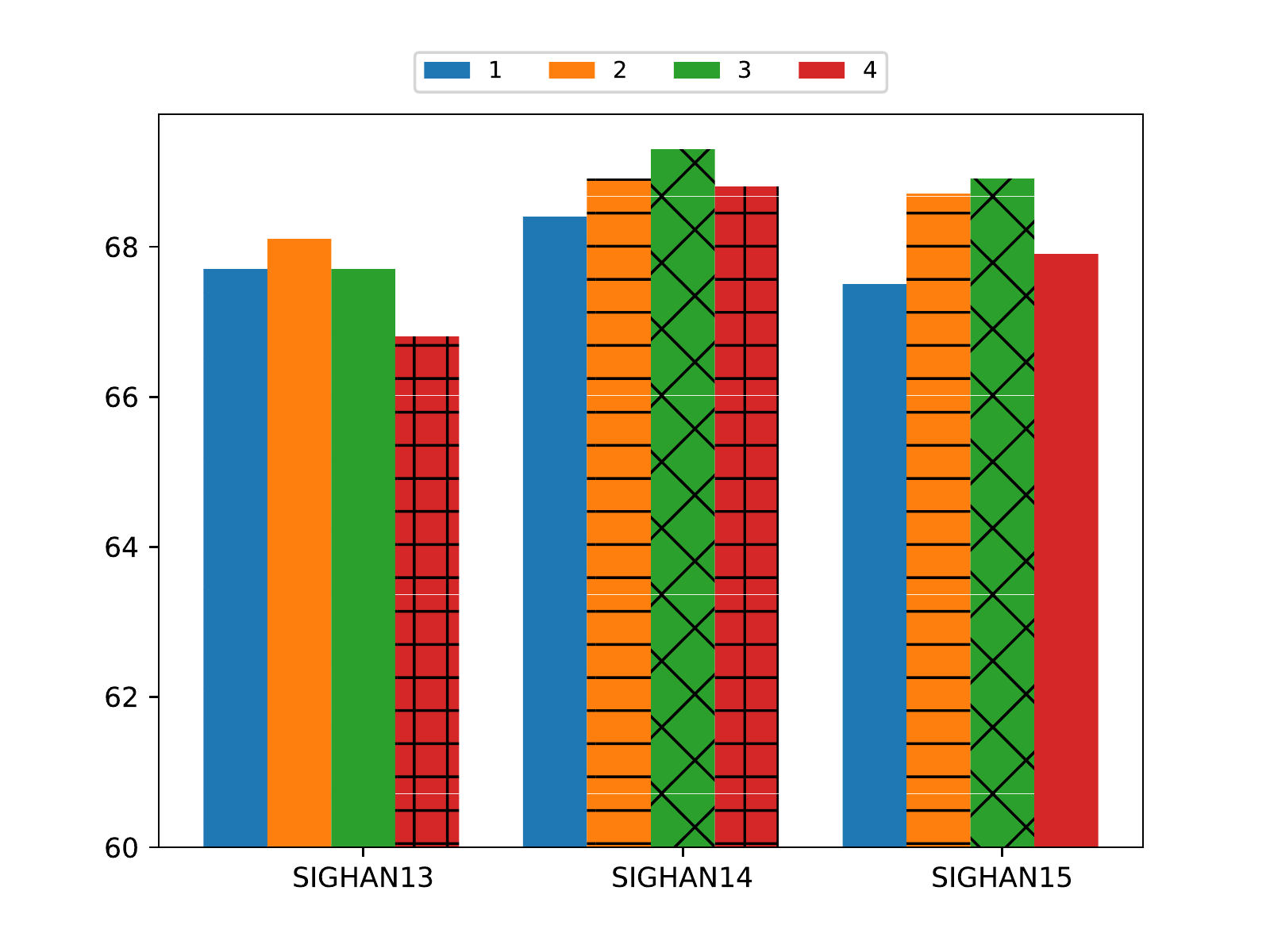}
\caption{The character-level C-F results (\%) w.r.t. the depth of SpellGCN.
The results were obtained with 10K training samples.
}
\label{fig:ablation-layers}
\end{figure}

\paragraph{Effect of the Number of Layers}
Generally, the performance of a GCN varies with the number of layers.
We investigate how the number of SpellGCN layers influence the performance in CSC.
In this comparison, the number of layers changes from 1 to 4, and the results are illustrated in Figure~\ref{fig:ablation-layers}.
For clarity, we report the character-level C-F on the three test datasets.
The results indicate that SpellGCN is able to make use of multiple layers.
With multiple layers, SpellGCN can aggregate the information in more hops and therefore, achieve better performance.
However, the F1score drops when the number of layers is larger than 3.
This is reasonable due to the over-smooth problem noted in~\cite{yan2019event}.
When the number of GCN layers increases, the representations of neighboring characters in the similarity graph will get more and more similar since they all are calculated via those of their neighbors in the similarity graph.

\paragraph{Effect of Attention Mechanism}
We investigate how to better combine the graphs in the SpellGCN layer.
Here, we compare the attention mechanism against sum-pooling and mean-pooling, with different hyper-parameter $\beta$ mentioned in Section~\ref{sec:attention}.
The experiments are conducted based on the 2-layer SpellGCN on SIGHAN 2013 test set.
The results presented in Table~\ref{tab:ablation-attention} show that the sum pooling fails in the CSC task. 
We suggest that the sum pooling is inconsistent with the normalization of GCN and fails to combine the information from different channels (i.e., graphs).
The mean pooling is feasible but is surpassed by the attention mechanism.
This indicates that the adaptive combination for each character node is beneficial.
We incorporate a hyper-parameter $\beta$ into the attention operation since the dot products may grow large in magnitude, pushing the softmax function into regions where it has extremely small gradients.
With these results, we chose the attention mechanism with a $\beta$ of 3 in SpellGCN.

\begin{table}[]
\centering
\small
\begin{tabular}{@{}l|r@{}}
\toprule
combination method                           & C-F         \\ \midrule
baseline (w/o SpellGCN)                                  & 67.0       \\ \midrule
sum pooling                                   & 66.3    \\
mean pooling                                  & 67.5       \\ \midrule
attentive combination ($\beta$=1)              & 67.8       \\
attentive combination ($\beta$=3)              & 68.2       \\
attentive combination ($\beta$=5)              & 68.0       \\
attentive combination ($\beta$=10)             & 67.7       \\ \bottomrule
\end{tabular}
\caption{The ablation results for graph combination method (\%).  
The averaged character-level C-F scores of 4 runs on the SIGHAN 2013 are reported.
The models were trained with 10K training samples.
Mean pooling denotes that the output representation $C^l$ of each layer is the average of $f_{k\in\{P,S\}}(\mathbf{A}_k,\mathbf{H}^l)$, while sum pooling summarizes $f_{k\in\{P,S\}}(\mathbf{A}_k,\mathbf{H}^l)$.
}
\label{tab:ablation-attention}
\end{table}

\subsection{Case Study}
We show several correction results to demonstrate the properties of SpellGCN.
In addition to the sample illustrated in Table~\ref{tab:example}, several prediction results are given in Table~\ref{tab:case}.
From these cases, we can tell that our SpellGCN is capable of revising the incorrect characters into correct ones with the pronunciation and shape constraint.
For instance, in the first case, ``麻坊(f\u{a}ng)'' is detected as errors and modified into ``麻烦(f\'an)''.
Without pronunciation similarity constraint, ``麻木(m\`u)'' becomes the most probable answer.
And surprisingly, in the second case, our SpellGCN successfully modifies the character reasonable in the context.
The meaning of input sentence ``看录音机'' is ``watch the audio recorder'', and our method corrects it into ``看录影机'' which means ``watch the video recorder''.
We suggest that SpellGCN injects a prior similarity between ``音'' and ``影'' in the representation space so that the model derives a higher posterior probability of ``影''.
In the last case, we show a correction result under the shape constraint.
In the confusion set, ``向'' is similar to ``尚'' and therefore, using SpellGCN is able to retrieve the correct result.

\begin{table}[]
\small
\centering
\resizebox{0.9\linewidth}{!}{%
\begin{tabular}{llll}
\toprule
\textit{Pronunciation: f\u{a}ng$\to$f\'an, w\`ang$\to$w\`ang} \\ \midrule
...走路真的麻{\color{orange}坊}，我也没有喝的东西，在家{\color{orange}汪}了...     \\
...走路真的麻{\color{orange}木}，我也没有喝的东西，在家{\color{orange}呆}了...     \\
...走路真的麻{\color{blue}烦}，我也没有喝的东西，在家{\color{blue}忘}了...     \\ \midrule
\textit{Pronunciation: y\=in$\to$y\u{i}ng} \\ \midrule
...因为妈妈或爸爸在看录{\color{orange}音}机...帮小孩子解决问题...  \\
...因为妈妈或爸爸在看录{\color{orange}音}机...帮小孩子解决问题...  \\
...因为妈妈或爸爸在看录{\color{blue}影}机...帮小孩子解决问题...  \\ \midrule
\textit{Shape: 向$\to$尚} \\ \midrule
...不过在许多传统文化的国家，女人{\color{orange}向}未得到平等...   \\
...不过在许多传统文化的国家，女人{\color{orange}从}未得到平等...   \\
...不过在许多传统文化的国家，女人{\color{blue}尚}未得到平等...   \\ \bottomrule
\end{tabular}
}
\caption{Several prediction results on the test set.
The first line in the block is the input sentence.
The second line is corrected by BERT without SpellGCN.
And the last line is the result from SpellGCN.
We highlight the {\color{orange}incorrect}/{\color{blue}correct} characters by {\color{orange}orange}/{\color{blue}blue} color.
}

\label{tab:case}
\end{table}

\subsection{Character Embedding Visualization}

Previous experiments have explored the performance of SpellGCN quantitatively in detail. 
To qualitatively study whether SpellGCN learns meaningful representations, we dive into the target  embedding space $\mathbf{W}$ derived from SpellGCN.

\begin{figure}[]
    \centering
    \begin{subfigure}[b]{0.48\linewidth}
        \centering
        \includegraphics[width=\linewidth]{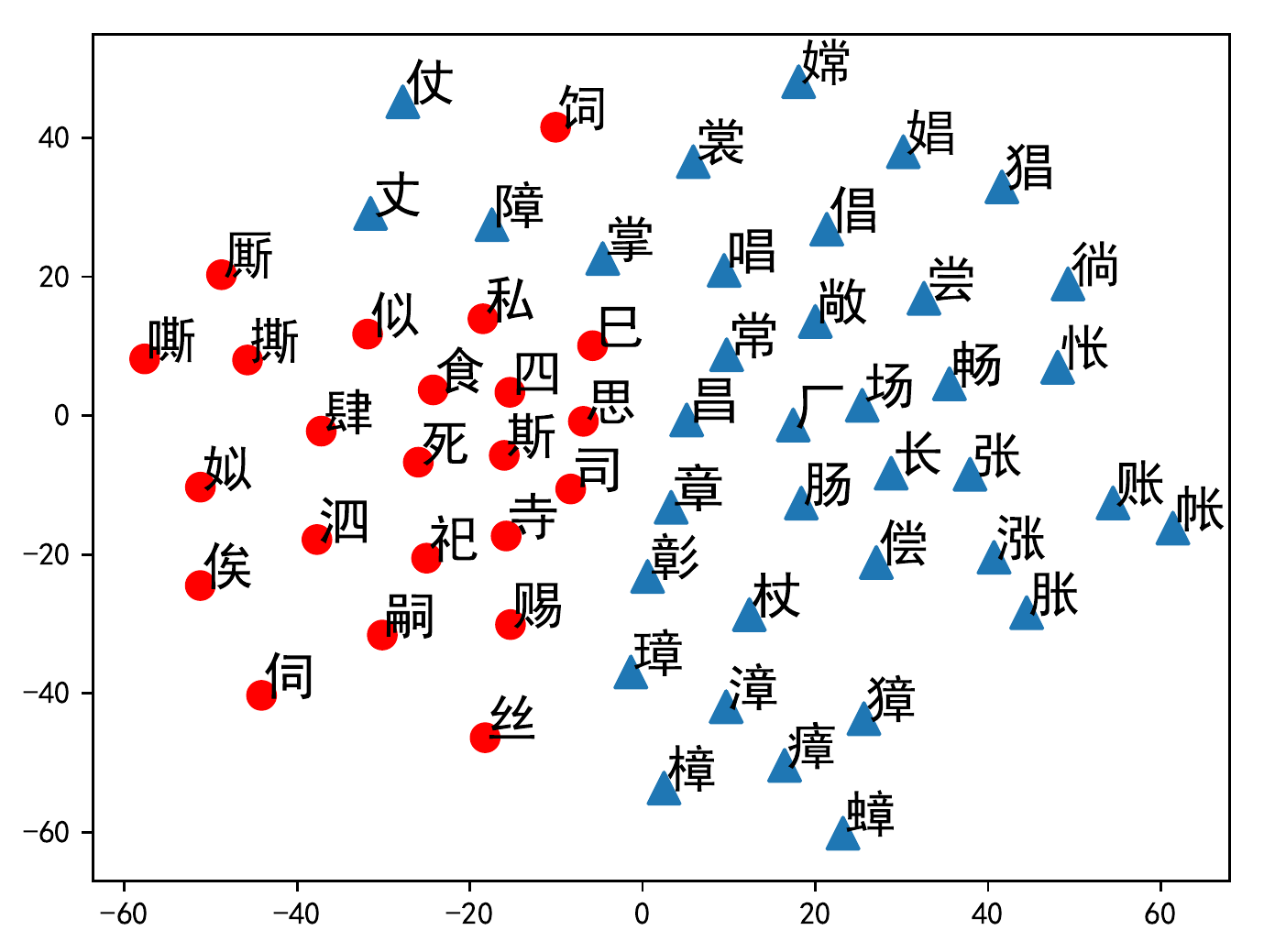}
        \caption{w/ SpellGCN}
    \end{subfigure}%
    \begin{subfigure}[b]{0.48\linewidth}
        \centering
        \includegraphics[width=\linewidth]{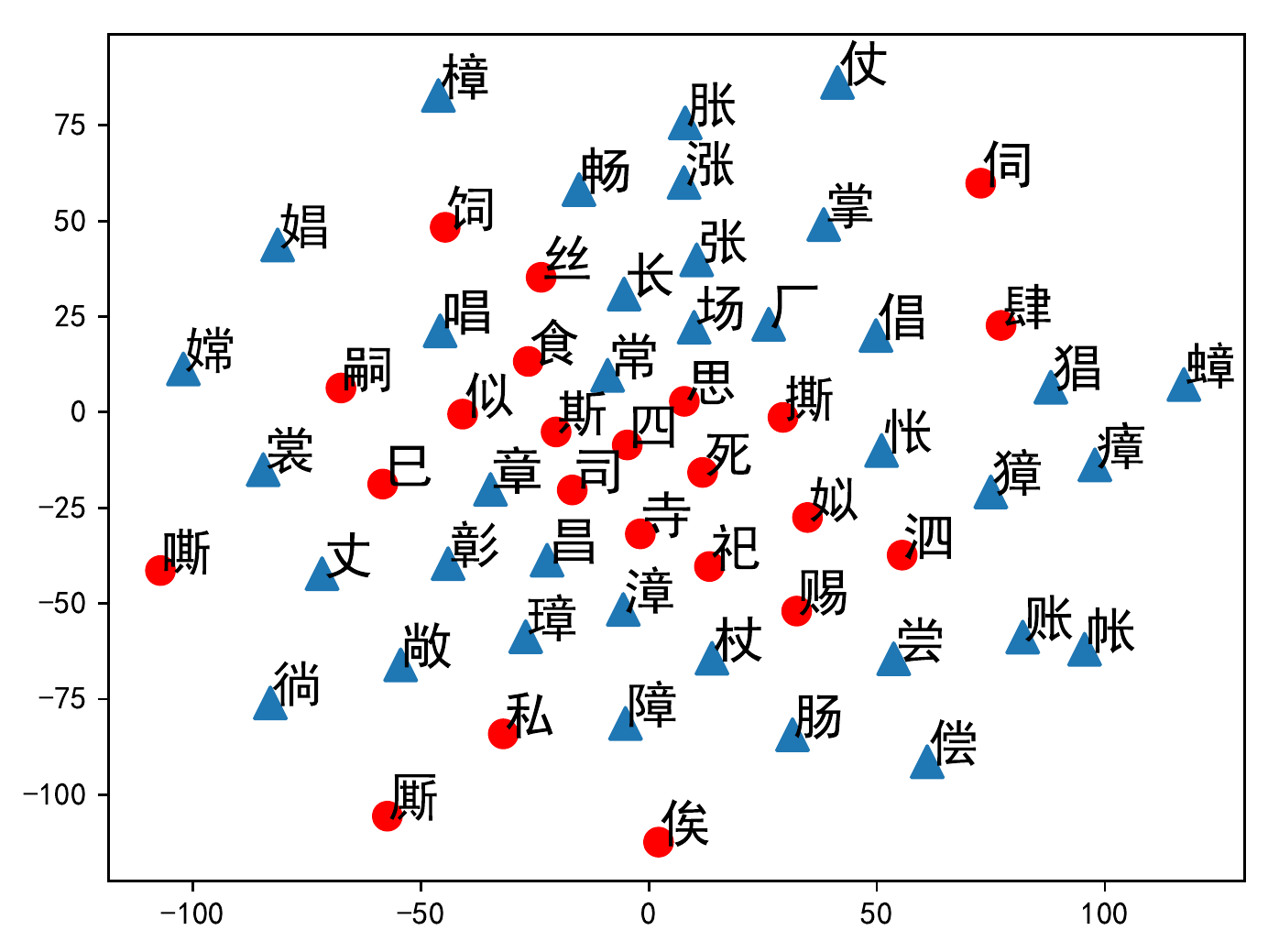}
        \caption{w/o SpellGCN}
    \end{subfigure}
    \caption{The scatter of similar characters of "长" and “祀” in terms of pronunciation by t-SNE.}
\label{fig:vis-pronunciation}
\end{figure}

\begin{figure}[]
\centering
    \begin{subfigure}[b]{0.48\linewidth}
        \centering
        \includegraphics[width=\linewidth]{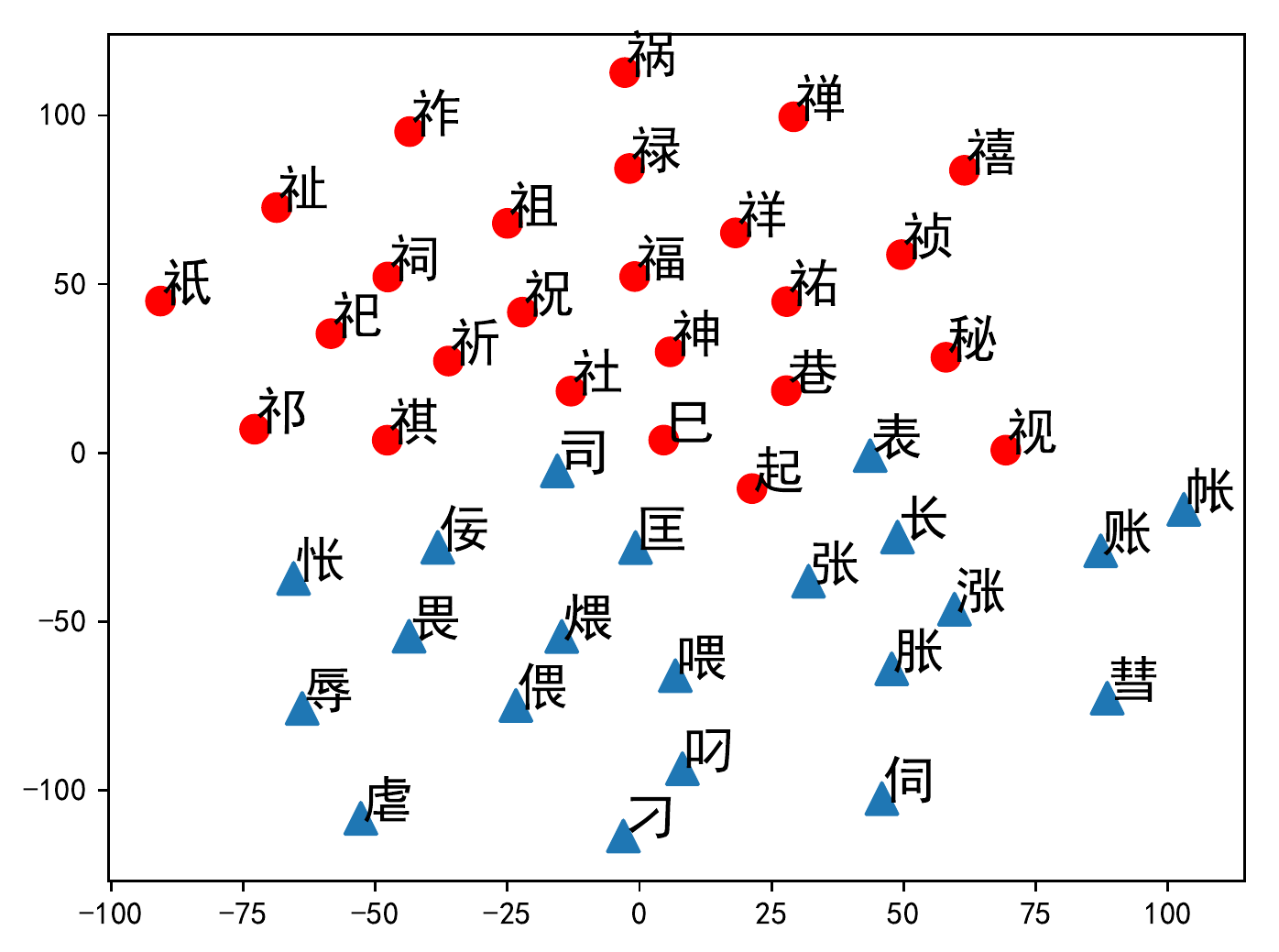}
        \caption{w/ SpellGCN}
    \end{subfigure}%
    \begin{subfigure}[b]{0.48\linewidth}
        \centering
        \includegraphics[width=\linewidth]{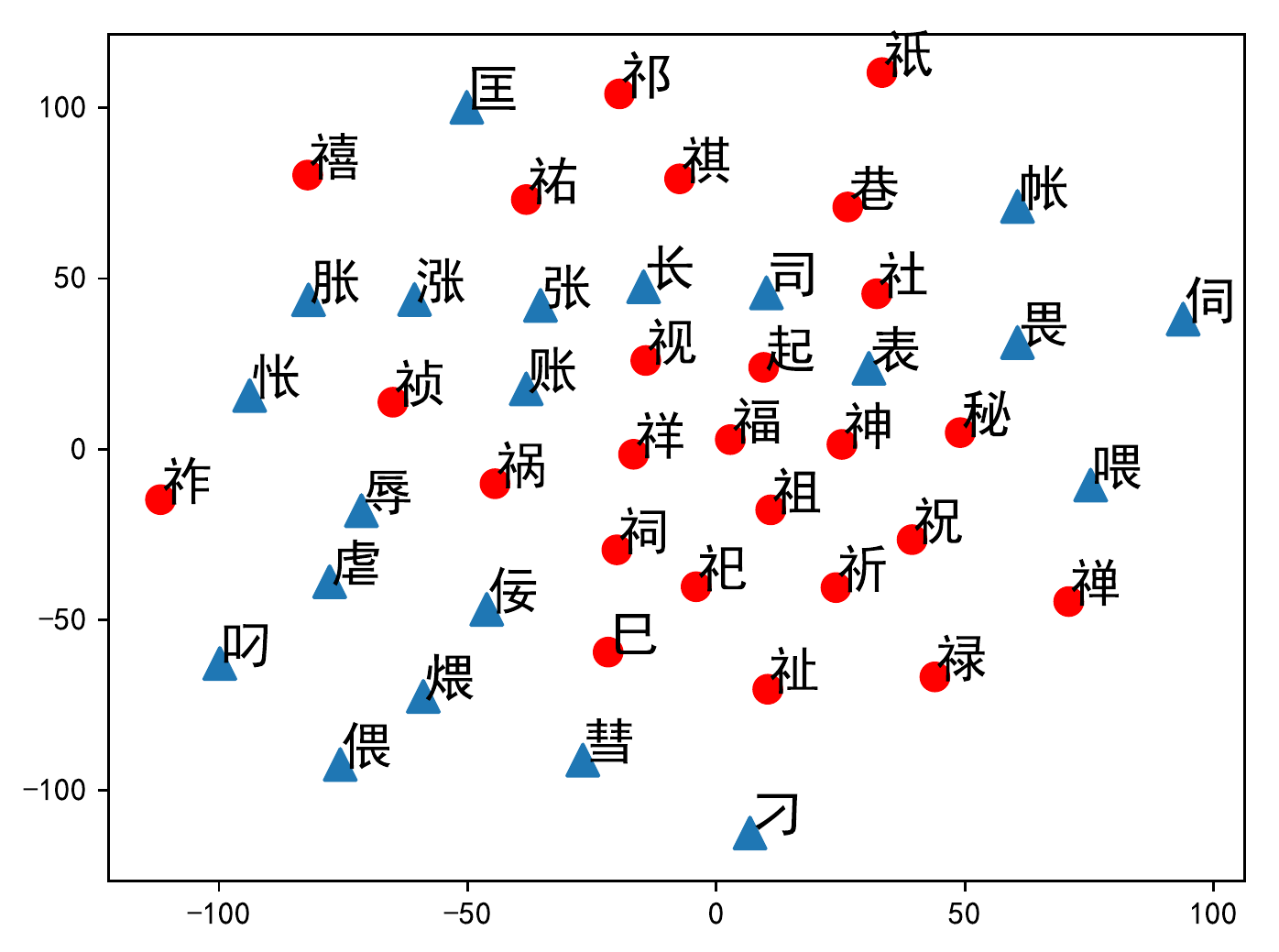}
       \caption{w/o SpellGCN}
    \end{subfigure}
    \caption{The scatter of similar characters of "长" and “祀” in terms of shape by t-SNE.}
\label{fig:vis-shape}
\end{figure}

In Figure~\ref{fig:vis-pronunciation}, the embedding of characters with phonics ``ch\'ang'' and ``s\`i'' is presented using t-SNE~\cite{maaten2008visualizing}.
The embedding learned by BERT captures the semantic similarity but fails to model the similarity in terms of pronunciation for the CSC task.
This is reasonable as this similarity knowledge is absent in the modeling.
In contrast, our SpellGCN successfully infuses this prior knowledge into the embedding and the resulting embedding exhibits cluster patterns.
The embedding of characters with these two different pronunciations forms two clusters, corresponding to ``ch\'ang'' and ``s\`i'' respectively.
Due to this property, the model tends to recognize similar characters and hence is able to retrieve the answers under pronunciation constraint.
Figure~\ref{fig:vis-shape} shows the same situation for the shape similarity, where two sets of characters with the shape similar to ``长'' and ``祀" are scattered.
This verifies the ability of SpellGCN in modeling shape similarity.

\section{Conclusions}
We proposed SpellGCN for CSC to incorporate both phonological and visual similarities into language models. 
The empirical comparison and the results of analytical experiments verify its effectiveness. 
Beyond CSC, SpellGCN can be generalized to other situations where specific prior knowledge is available, and to other languages by leveraging specific similarity graphs analogously. 
Our method can also be adapted to grammar error correction, which needs insertion and deletion, by utilizing more flexible extractors such as Levenshtein Transformer~\cite{DBLP:journals/corr/abs-1905-11006}.
We leave this direction to future work.



\bibliography{anthology,acl2020}
\bibliographystyle{acl_natbib}

\appendix


\end{CJK*}
\end{document}


\maketitle
This document is to present several additional results in complementary to the main paper.
Although these results are not directly related to the content in the paper, they are worth-noting and may shed light on following research.

\section{SpellBERT}
In our early experiments, we proposed a modified masking strategy of BERT for CSC, to which we refer to SpellBERT.
Briefly, the SpellBERT is useful when used as the extractor with small training data, i.e., without 271K additional generated data.
However, it is not beneficial with full training data.

The motivation of SpellBERT is straightforward: although SpellGCN is able to infuse the similarity knowledge into the model in the fine-tuning phase, it would be better if this knowledge is inherent.
For this goal, we changed the language mask strategy when pre-training a BERT.
In the original mask strategy of BERT, the masked character will be replaced by a \texttt{[MASK]} token 80\% of the time, or a random token from the vocabulary 10\% of the time, or unchanged token 10\% of the time.
In this work, we include a masking strategy that selects a random token from the confusion set.
The BERT with modified masking strategy is called SpellBERT.

The results of SpellBERT with 10K training data are given in Table~\ref{tab:spellbert}.
From these results, SpellBERT is able to obtain additional improvement over the vanilla BERT.
And this improvement is compatible with SpellGCN.
The combination of SpellBERT and SpellGCN trained on 10K training data outperforms the PN model~\cite{pn19} using 280K training data.
Among various masking strategies, a split of (80\%, 6.6\%, 6.7\%, 6.7\%) achieves best performance, cf. Table~\ref{tab:ablation-spellbert}.

\begin{table}[]
\centering
\resizebox{\textwidth}{!}{%
\begin{tabular}{@{}l|lll|lll|lll@{}}
\toprule
SIGHAN13              & D-P    & D-R    & D-F    & C-P    & C-R    & C-F    & J-P    & J-R    & J-F    \\ \midrule
GoogleBERT + SpellGCN & 69.7\% & 68.9\% & 69.3\% & 83.5\% & 57.5\% & 68.1\% & 69.7\% & 57.5\% & 63.0\% \\
SpellBERT + SpellGCN  & 69.3\% & 72.9\% & 71.0\% & 82.0\% & 73.1\% & 77.3\% & 68.5\% & 63.0\% & 65.7\% \\
\bottomrule \toprule
SIGHAN14              & D-P    & D-R    & D-F    & C-P    & C-R    & C-F    & J-P    & J-R    & J-F    \\ \midrule
GooglBERT + SpellGCN  & 74.8\% & 65.0\% & 69.6\% & 86.8\% & 56.5\% & 68.4\% & 77.5\% & 56.5\% & 65.3\% \\
SpellBERT + SpellGCN  & 74.6\% & 67.0\% & 70.6\% & 84.2\% & 73.0\% & 78.2\% & 78.3\% & 60.5\% & 68.3\% \\
\bottomrule \toprule
SIGHAN15              & D-P    & D-R    & D-F    & C-P    & C-R    & C-F    & J-P    & J-R    & J-F    \\ \midrule
GooglBERT + SpellGCN  & 75.8\% & 70.5\% & 73.0\% & 82.8\% & 58.4\% & 68.5\% & 75.7\% & 58.4\% & 65.9\% \\
SpellBERT + SpellGCN  & 75.0\% & 69.1\% & 71.9\% & 81.0\% & 72.8\% & 76.7\% & 78.6\% & 60.3\% & 68.2\% \\ \bottomrule
\end{tabular}%
}
\caption{The comparison between SpellBERT and BERT (Google) trained using 10K training data.
D, C, J denote the detection, correction and the joint one.
}
\label{tab:spellbert}
\end{table}

\begin{table}[]
\centering
\resizebox{\textwidth}{!}{%
\begin{tabular}{@{}l|lll|lll|lll@{}}
\toprule
SIGHAN13                                 & D-P    & D-R    & D-F    & C-P    & C-R    & C-F    & J-P    & J-R    & J-F    \\ \midrule
SpellBERT (80.0, 06.6, 06.7, 06.7, 00.0) & 50.7\% & 53.9\% & 52.2\% & 82.5\% & 44.5\% & 57.8\% & 52.1\% & 44.5\% & 48.0\% \\
SpellBERT (80.0, 10.0, 10.0, 00.0, 00.0) & 47.4\% & 49.5\% & 48.4\% & 73.0\% & 36.1\% & 48.3\% & 44.4\% & 36.1\% & 39.8\% \\
SpellBERT (80.0, 06.6, 00.0, 06.7, 06.7) & 49.2\% & 51.5\% & 50.3\% & 82.8\% & 42.7\% & 56.3\% & 51.7\% & 42.7\% & 46.7\% \\
SpellBERT (00.0, 33.3, 00.0, 33.4, 33.4) & 49.8\% & 50.8\% & 50.3\% & 83.9\% & 42.6\% & 56.5\% & 54.6\% & 42.6\% & 47.9\% \\ 
\bottomrule \toprule
SIGHAN14                                 & D-P    & D-R    & D-F    & C-P    & C-R    & C-F    & J-P    & J-R    & J-F    \\ \midrule
SpellBERT (80.0, 06.6, 06.7, 06.7, 00.0) & 67.2\% & 54.8\% & 60.4\% & 88.7\% & 48.6\% & 62.8\% & 78.7\% & 48.6\% & 60.1\% \\
SpellBERT (80.0, 10.0, 10.0, 00.0, 00.0) & 63.6\% & 52.3\% & 57.4\% & 81.8\% & 42.8\% & 56.2\% & 70.1\% & 42.8\% & 53.1\% \\
SpellBERT (80.0, 06.6, 00.0, 06.7, 06.7) & 64.3\% & 53.2\% & 58.2\% & 89.8\% & 47.7\% & 62.3\% & 78.4\% & 47.7\% & 59.3\% \\
SpellBERT (00.0, 33.3, 00.0, 33.4, 33.4) & 61.9\% & 50.3\% & 55.5\% & 88.7\% & 44.6\% & 59.3\% & 76.5\% & 44.6\% & 56.3\% \\
\bottomrule \toprule
SIGHAN15                                 & D-P    & D-R    & D-F    & C-P    & C-R    & C-F    & J-P    & J-R    & J-F    \\ \midrule
SpellBERT (80.0, 06.6, 06.7, 06.7, 00.0) & 65.7\% & 58.5\% & 61.9\% & 82.5\% & 48.3\% & 60.9\% & 73.1\% & 48.3\% & 58.1\% \\
SpellBERT (80.0, 10.0, 10.0, 00.0, 00.0) & 64.7\% & 56.5\% & 60.3\% & 79.0\% & 44.6\% & 57.0\% & 69.2\% & 44.6\% & 54.3\% \\
SpellBERT (80.0, 06.6, 00.0, 06.7, 06.7) & 66.5\% & 58.2\% & 62.0\% & 82.0\% & 47.7\% & 60.3\% & 72.4\% & 47.7\% & 57.5\% \\
SpellBERT (00.0, 33.3, 00.0, 33.4, 33.4) & 66.2\% & 57.9\% & 61.8\% & 81.4\% & 47.1\% & 59.7\% & 73.7\% & 47.1\% & 57.5\% \\ \bottomrule
\end{tabular}%
}
\caption{Ablation studies for SpellBERT in terms of masking strategy.
The numbers in the model name are ((1) \texttt{[MASK]} token, (2) a random token from the vocabulary, (3) unchanged token, (4) a similar token from the confusion set, (5) a random token from the confusion set). 
D, C, J denote the detection, correction and the joint one.
Due to computational resource limitation, these SpellBERT are 6-layer version.
}
\label{tab:ablation-spellbert}
\end{table}

However, we observed opposite results when additional 271K training data is used, as illustrated in Table~\ref{tab:spellbert-large}.
SpellBERT was surpassed by vanilla BERT.
Regard this negative result, we suggest that SpellBERT is able to learn the similarity information from the specified masking strategies, but its ability in semantic reasoning is harmed.
With large amount of training data, the semantic ability becomes essential.

\begin{table}[]
\centering
\resizebox{\textwidth}{!}{%
\begin{tabular}{l|lll|lll|lll}
\toprule
SIGHAN13              & D-P    & D-R    & D-F    & C-P    & C-R    & C-F    & J-P    & J-R    & J-F    \\ \midrule
GoogleBERT + SpellGCN & 74.7\% & 90.0\% & 81.6\% & 98.5\% & 88.7\% & 93.4\% & 76.4\% & 88.7\% & 82.1\% \\
SpellBERT + SpellGCN  & 74.1\% & 88.7\% & 80.7\% & 98.3\% & 87.2\% & 92.4\% & 75.8\% & 87.2\% & 81.1\% \\ 
\bottomrule \toprule
SIGHAN14              & D-P    & D-R    & D-F    & C-P    & C-R    & C-F    & J-P    & J-R    & J-F    \\ \midrule
GooglBERT + SpellGCN  & 83.6\% & 78.6\% & 81.0\% & 97.2\% & 76.4\% & 85.5\% & 88.0\% & 77.1\% & 82.2\% \\
SpellBERT + SpellGCN  & 80.3\% & 77.6\% & 78.9\% & 96.7\% & 75.1\% & 84.5\% & 83.8\% & 75.1\% & 79.2\% \\\bottomrule \toprule
SIGHAN15              & D-P    & D-R    & D-F    & C-P    & C-R    & C-F    & J-P    & J-R    & J-F    \\ \midrule
GooglBERT + SpellGCN  & 88.9\% & 87.7\% & 88.3\% & 95.7\% & 83.9\% & 89.4\% & 89.2\% & 83.9\% & 86.4\% \\ 
SpellBERT + SpellGCN  & 84.7\% & 83.7\% & 84.2\% & 95.7\% & 80.1\% & 87.2\% & 86.9\% & 80.1\% & 83.4\% \\
\bottomrule
\end{tabular}%
}
\caption{The comparison between SpellBERT and BERT (Google) trained using 280K training data.
D, C, J denote the detection, correction and the joint one.
The results of SIGHAN13 are different from the results of Table 3 in the paper as this model is not further fine-tuned.
}
\label{tab:spellbert-large}
\end{table}

We were also interested in using SpellGCN in pre-training the BERT.
However, the resulting model failed.
This indicates that pre-training is crucial and plays a role of semantic reasoning for downstream tasks.
And it is detrimental to learn the strong similarity regularization in the pre-training.

\bibliography{anthology,acl2020}
\bibliographystyle{acl_natbib}